\documentclass[letterpaper]{article} 
\usepackage{aaai2026}  
\usepackage{times}  
\usepackage{helvet}  
\usepackage{courier}  
\usepackage[hyphens]{url}  
\usepackage{graphicx} 
\usepackage{natbib}  
\usepackage{caption} 
\usepackage{algorithm}
\usepackage{algorithmic}

\usepackage{amssymb}
\usepackage{amsmath}
\usepackage{booktabs}
\usepackage{multirow}
\usepackage{pifont}

\usepackage{newfloat}
\usepackage{listings}
\DeclareCaptionStyle{ruled}{labelfont=normalfont,labelsep=colon,strut=off} 
\floatstyle{ruled}
\newfloat{listing}{tb}{lst}{}
\floatname{listing}{Listing}
\title{A Unified Shape-Aware Foundation Model for Time Series Classification}

\author {
    Zhen Liu\textsuperscript{\rm 1,\rm 2},
    Yucheng Wang\textsuperscript{\rm 2},
    Boyuan Li\textsuperscript{\rm 1},
    Junhao Zheng\textsuperscript{\rm 1},
    Emadeldeen Eldele\textsuperscript{\rm 3}, \\
    Min Wu\textsuperscript{\rm 2}$^*$,
    Qianli Ma\textsuperscript{\rm 1}\thanks{Corresponding authors.}
}
\affiliations {
    \textsuperscript{\rm 1}School of Computer Science and Engineering, South China University of Technology, Guangzhou, China\\
    \textsuperscript{\rm 2}Institute for Infocomm Research, Agency for Science, Technology and Research, Singapore\\
     \textsuperscript{\rm 3}Department of Computer Science, Khalifa University, UAE\\
    cszhenliu@foxmail.com, wumin@a-star.edu.sg, qianlima@scut.edu.cn
}

\usepackage{bibentry}

\begin{document}

\maketitle

\begin{abstract}
Foundation models pre-trained on large-scale source datasets are reshaping the traditional training paradigm for time series classification.
However, existing time series foundation models primarily focus on forecasting tasks and often overlook classification-specific challenges, such as modeling interpretable shapelets that capture class-discriminative temporal features. 
To bridge this gap,  we propose UniShape, a unified shape-aware foundation model designed for time series classification. 
UniShape incorporates a shape-aware adapter that adaptively aggregates multiscale discriminative subsequences (shapes) into class tokens,  effectively selecting the most relevant subsequence scales to enhance model interpretability. Meanwhile, a prototype-based pretraining module is introduced to jointly learn instance- and shape-level representations, enabling the capture of transferable shape patterns. 
Pre-trained on a large-scale multi-domain time series dataset comprising 1.89 million samples, UniShape exhibits superior generalization across diverse target domains.
Experiments on 128 UCR datasets and 30 additional time series datasets demonstrate that UniShape achieves state-of-the-art classification performance, with interpretability and ablation analyses further validating its effectiveness. 
\end{abstract}

\begin{links}
    \link{Code, Datasets, and Supplementary Material}{https://github.com/qianlima-lab/UniShape}
\end{links}

\section{Introduction}

Deep learning models have achieved notable success in time series classification (TSC) across various domains~\cite{ismail2019deep,luo2024knowledge}.
However, most existing methods~\cite{liu2023scale,mohammadi2024deep} are trained on small-scale datasets, thereby limiting their generalization capability in cross-domain settings.
In contrast, foundation models (FMs) have exhibited strong transferability in vision and language tasks~\cite{zhang2024vision}, prompting a growing interest in their application to time series data. Yet, existing efforts mainly focused on time series forecasting tasks~\cite{ansari2024, li2025tsfm}, while FM development specifically for classification tasks is still in its early stages.  Designing a unified FM for TSC thus presents an open and impactful research problem.

\begin{figure}
	\centering 
    \includegraphics[ width=0.44\textwidth]{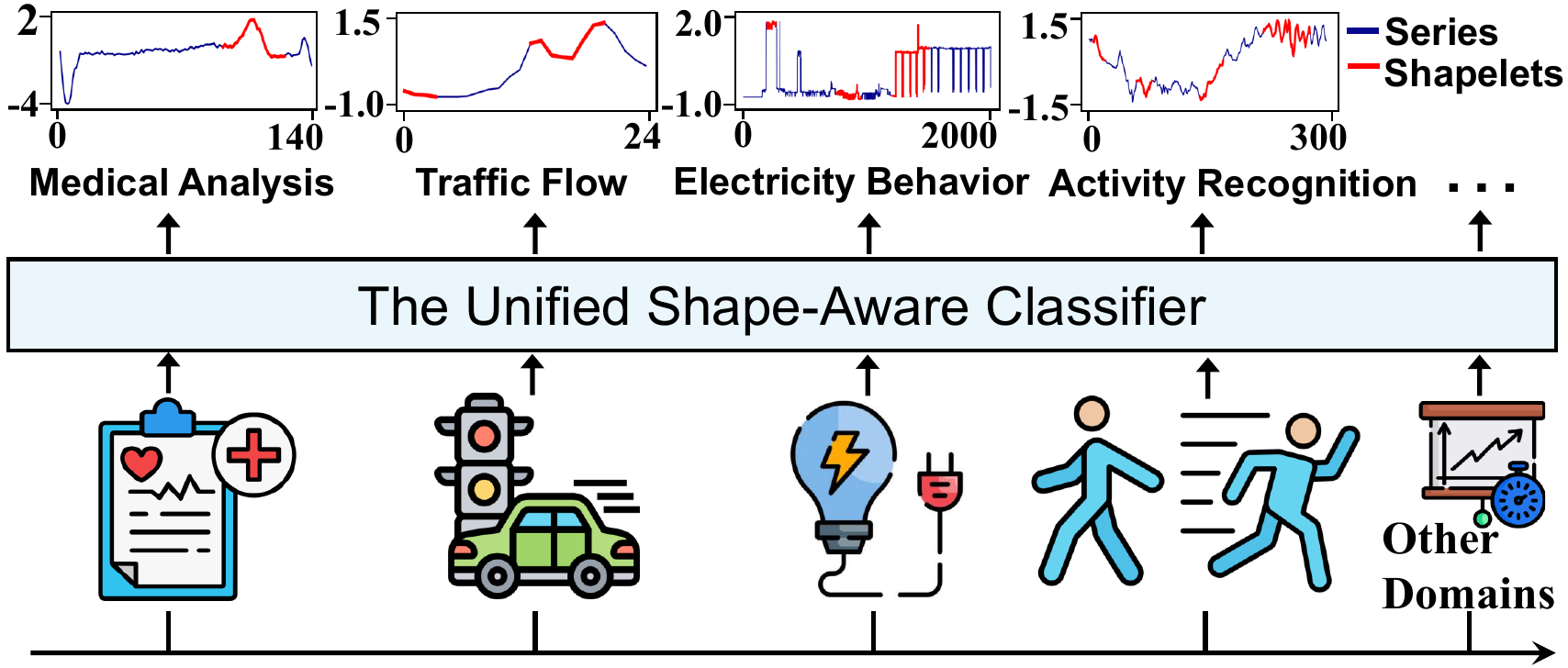}
	\caption{
Illustration of the unified shape-aware classifier trained across multiple domains.
Red lines represent shapelets extracted from the \textit{ECG5000}, \textit{Chinatown}, \textit{HouseTwenty}, and \textit{CricketX} datasets, corresponding to four distinct domains in the UCR time series archive~\cite{dau2019ucr}.
    } 
	\label{fig:motivation_vis}
\end{figure}

The fundamental differences between time series forecasting and classification present key challenges for developing effective FMs for classification. Forecasting focuses on learning temporal dynamics such as trends and seasonality from historical data~\cite{woocost2022}, aiming to predict future values based on sufficiently long contextual input sequences. Conversely, classification emphasizes identifying discriminative local patterns within fixed-length samples from different subjects~\cite{ismail2020inceptiontime,liu2023temporal}, thus assigning discrete class labels to unseen samples. While forecasting outputs continuous, multi-step numerical sequences, classification involves extracting informative and interpretable patterns from individual samples. 
As a result, forecasting-oriented FMs often fail to capture class-discriminative features essential for accurate TSC.

Although a few recent studies~\cite{goswami2024moment,feofanov2025mantis} have made initial attempts to apply FMs for TSC, they largely neglect the interpretability that is vital to domains such as healthcare.
Shapelets are discriminative subsequences widely used to enhance the interpretability of TSC results, as they can represent key class patterns (e.g., the red segment in Figure~\ref{fig:motivation_vis})~\cite{ye2011time}.
However, existing shapelet-based methods typically rely on label supervision with domain-specific assumptions~\cite{li2021shapenet, liusoftshape2025}, limiting their applicability in FM pretraining scenarios where annotations are scarce.
Also, shapelets inherently exhibit multiscale properties~\cite{grabocka2014learning, yamaguchi2023time}, as discriminative subsequence patterns may appear at varying lengths and temporal locations. Therefore, effectively modeling and integrating multiscale shapelet representations within a unified FM during pretraining remains a challenge.

To this end, this paper proposes a \textbf{Uni}fied \textbf{Shape}-aware foundation model named \textbf{UniShape} for TSC that aims to enhance both downstream classification performance and interpretability. 
UniShape comprises two key components.  First, the shape-aware adapter processes multiscale subsequences (or shapes) as input tokens and adaptively aggregates discriminative features of varying lengths via an attention-based pooling mechanism to generate instance-level class tokens. 
Building on these class and shape tokens, 
the prototype-based pretraining module then jointly optimizes instance-level and shape-level representations for class prototype learning, enabling the model to capture generalizable shapelet patterns.  Through this design, UniShape achieves effective transferability and improved interpretability when fine-tuned on diverse TSC target domains following pretraining on a large-scale multi-domain dataset.

The main contributions are summarized as follows:

\begin{itemize}

\item  We propose UniShape, a unified foundation model for TSC that effectively captures multiscale shapelet features and automatically selects optimal subsequence scales via a shape-aware adapter.

\item  We introduce a prototype-based pretraining module that jointly learns discriminative representations at both instance-level and shape-level tokens, thus enhancing the model's generalization capability in target domains.

\item  Extensive experiments on 158 univariate time series datasets demonstrate that UniShape significantly outperforms state-of-the-art methods in classification performance, while also exhibiting good interpretability.

\end{itemize}

\section{Related Work}

\subsubsection{Time Series Classification.}

Early TSC approaches primarily relied on dynamic time warping and nearest neighbor classifiers~\cite{keogh2002need}. Recent efforts have shifted towards non-deep learning methods~\cite{guillaume2022random, middlehurst2024bake}, such as the Rocket family of algorithms~\cite{dempster2020rocket, dempster2021minirocket, dempster2023hydra}, alongside deep learning models based on neural networks~\cite{mohammadi2024deep}. In particular, both self-supervised representation learning~\cite{ yue2022ts2vec} and end-to-end supervised classification frameworks~\cite{ismail2020inceptiontime,eldele2024tslanet} have shown strong performance. However, most existing TSC models are trained on single-domain datasets, limiting their transferability across multiple domains.

\subsubsection{Time Series Foundation Models.}

Time series FMs have seen rapid development~\cite{liang2024foundation}. For example, recent studies exhibit strong forecasting generalization by pretraining on large-scale datasets merged from multiple domains~\cite{woo2024unified,li2025tsfm}.
\citet{zhou2023one} further demonstrate the adaptability of large language models, while \citet{goswami2024moment} and \citet{gao2024units} design task-agnostic FMs pretrained on time series datasets for multiple downstream tasks (including classification).
For classification-specific FMs, ~\citet{linnutime2024} and~\citet{feofanov2025mantis} introduce multi-scale normalization techniques, highlighting the potential of FMs for TSC. However, most efforts focus on enhancing classification accuracy, with limited attention to the interpretability of the results.

\subsubsection{Time Series Shapelets.}

Shapelets have attracted considerable interest for enhancing the interpretability of TSC models~\cite{liu2024diffusion, wen2024abstracted}. Traditional approaches~\cite{ye2011time,rakthanmanon2013fast} rely on exhaustive searches to identify discriminative subsequences, incurring high computational costs. Recent deep learning methods~\cite{li2021shapenet,liusoftshape2025} adopt gradient-based frameworks to improve shapelet discovery efficiency.  
However, most existing methods are restricted to scenario-specific shapelet discovery, limiting their applicability as universal representations across diverse domains.

\section{Background}

\subsubsection{Problem Statement.}

This paper focuses on building a foundation model for the univariate TSC problem, using a two-stage paradigm: pretraining and fine-tuning. 
The model is first pre-trained on a large-scale source dataset and then fine-tuned on domain-specific target datasets.  Compared to training deep learning models from scratch, FMs provide better parameter initialization and enhanced generalization~\cite{matkde2024}, making them particularly suitable for transfer learning across multiple target domains. 

Formally, let the pretraining source dataset be denoted as
$
\mathcal{D}_s = \{(\mathbf{x}^{(i)}, y^{(i)})\}_{i=1}^N,
$
where each time series sample
$
\mathbf{x}^{(i)} = [x_1^{(i)}, x_2^{(i)}, \dots, x_T^{(i)}] \in \mathbb{R}^T
$
has length $T$, and $y^{(i)}$ denotes its class label. 
A foundation model $f(\cdot)$ is first pre-trained on $\mathcal{D}_s$, 
and its parameters are then frozen and used to initialize downstream training.
During fine-tuning, the model $f(\cdot)$ is further trained on the target dataset   
$
\mathcal{D}_t = \{(\mathbf{x}_t^{(j)}, y_t^{(j)})\}_{j=1}^M,
$
where $\mathbf{x}_t^{(j)} \in \mathbb{R}^T$ and $y_t^{(j)}$ represent the input data and label in the target dataset. The goal is to use pre-trained knowledge learned from $\mathcal{D}_s$ to improve classification performance on the target dataset $\mathcal{D}_t$.

\begin{figure*}[!thb]
\vspace{1em}
	\centering 
	\includegraphics[ width=0.89\textwidth]{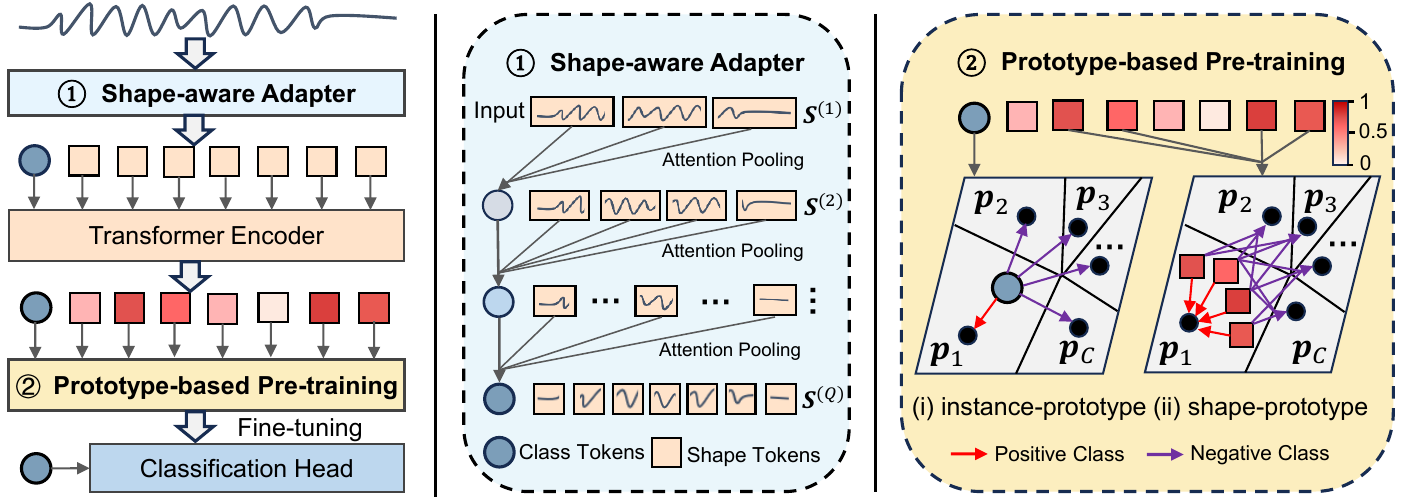}
    \vspace{1em}
  \caption{
 The overall architecture of the UniShape framework.
UniShape comprises two core modules: {\Large \ding{172}} a shape-aware adapter that takes variable-length subsequences $[\mathcal{S}^{(q)}]_{q=1}^{Q}$ as input and applies attention pooling to fuse discriminative patterns into class tokens; and {\Large \ding{173}} a prototype-based pretraining module that jointly uses instance-prototype and shape-prototype contrastive learning to optimize prototype representations $\{\mathbf{p}_c\}_{c=1}^{C}$ based on instance-level class tokens and subsequence-level shape tokens.
  }
    \vspace{0.5em}
	\label{fig:uni_shape_framework}
\end{figure*}

\subsubsection{The Pretraining Source Dataset.}
A large-scale time series source dataset is essential for effective foundation model pretraining.
However, most existing TSC datasets are small and domain-specific, limiting their utility for this purpose.  To address this limitation, we construct a comprehensive pretraining dataset following~\cite{linnutime2024}, integrating three primary sources: (1) the UCR time series archive~\cite{dau2019ucr}, (2) the UEA time series archive~\cite{bagnall2018uea}, and (3) eight additional datasets commonly used in prior studies~\cite{eldele2021time,zhang2022self}.

To ensure consistency in input channels across domains, multivariate sequences are decomposed into distinct univariate series using a channel-independent transformation. To address inconsistencies in sequence length, all inputs are resized to a fixed length of 512 using PyTorch's interpolation function~\cite{feofanov2025mantis}. To prevent test data leakage, only the training sets of each sub-dataset from the above three data sources are employed. The resulting source dataset contains approximately 1.89 million univariate time series samples, providing a large-scale and diverse corpus suitable for time series foundation model pretraining.

\section{Method}

\subsection{Architecture Overview}

As shown in Figure~\ref{fig:uni_shape_framework}, the overall architecture of UniShape comprises two key components: (i) a shape-aware adapter and (ii) a prototype-based pretraining module.

The shape-aware adapter begins by segmenting each input time series into multi-scale subsequences (or shapes) using sliding windows of varying lengths. Each group of fixed-length subsequences is processed by a shared-parameter adapter that aggregates shape tokens into a class token via attention pooling. These class tokens are then hierarchically fused in a top-down manner to capture discriminative representations of shapes across scales.

In the prototype-based pretraining module, the class and shape tokens generated by the adapter are fed into a transformer encoder. Class prototype learning is performed at two levels: instance-level, utilizing class tokens to capture global class-discriminative features across samples; and shape-level, selecting high-confidence shape tokens via class attention scores to model local discriminative patterns within individual samples.
After pretraining, UniShape fine-tunes on target datasets by passing the final class token into a classification head, yielding the classification output.

\subsection{Shape-Aware Adapter}

Deep learning models typically require fixed-length inputs, posing challenges for handling variable-length time series with different sampling scales. A naive solution, which involves training separate models for each input length, is inefficient and hinders transferability in FM design.
Inspired by the success of adapters in NLP for their parameter efficiency and scalability~\cite{houlsby2019parameter}, we propose a \emph{shape-aware adapter} to enable unified and efficient modeling of multi-scale subsequences within time series FMs.

Given a univariate time series $\mathbf{x} = [x_1, x_2, \dots, x_T] \in \mathbb{R}^T$, we define $Q$ temporal scales using sliding window configurations $(W_q, K_q)$ for $q \in [1, Q]$, where $W_q$ and $K_q$ denote the window length and stride, respectively. Each scale $q$ produces a subsequence set:
\begin{equation}
\mathcal{S}^{(q)} = \left\{ \mathbf{s}^{(q)}_i = [x_i, \dots, x_{i+W_q - 1}] \in \mathbb{R}^{W_q} \right\},
\label{eq:subsequence_set}
\end{equation}
where $i = 1, 1+K_q, \dots, T-K_q$. This yields $N_q = \left\lfloor \frac{T - W_q}{K_q} \right\rfloor + 1$ subsequences at scale $q$.

Each subsequence $\mathbf{s}^{(q)}_i$ is transformed into a $d$-dimensional shape token $\mathbf{z}^{(q)}_i$ via a lightweight normalization unit~\cite{feofanov2025mantis}. To obtain $\mathbf{z}^{(q)}_i$, we first compute the mean $\mu$ and standard deviation $\sigma$ of $\mathbf{x}$, as well as its first-order differential $\Delta \mathbf{x} = [x_2 - x_1, \dots, x_T - x_{T-1}, 0]$. Each subsequence and its differential form are normalized:
$
\hat{\mathbf{s}}^{(q)}_i = \frac{\mathbf{s}^{(q)}_i - \mu}{\sigma}, \quad 
\widehat{\Delta \mathbf{s}}^{(q)}_i = \frac{\Delta \mathbf{s}^{(q)}_i - \Delta\mu}{\Delta\sigma}.
$
Then, two 1D CNNs encode these normalized inputs:
$
\mathbf{h}^{(q)}_i = \text{CNN}_1(\hat{\mathbf{s}}^{(q)}_i), \quad 
\mathbf{g}^{(q)}_i = \text{CNN}_2(\widehat{\Delta \mathbf{s}}^{(q)}_i).
$
In parallel, local mean $\mu^{(q)}_i$ and standard deviation $\sigma^{(q)}_i$ of each subsequence $\mathbf{s}^{(q)}_i$ are embedded using a numerically multi-scaled embedding module~\cite{linnutime2024}, yielding:
$\mathbf{e}(\mu^{(q)}_i), \mathbf{e}(\sigma^{(q)}_i) \in \mathbb{R}^d$.
Finally, all embeddings are concatenated and linearly projected into a shape token:
\begin{equation}
\mathbf{z}^{(q)}_i = \text{Linear}\left( 
\left[ \mathbf{h}^{(q)}_i, \mathbf{g}^{(q)}_i, \mathbf{e}(\mu^{(q)}_i), \mathbf{e}(\sigma^{(q)}_i) \right] 
\right) \in \mathbb{R}^d.
\label{eq:shape_token_embedding}
\end{equation}

To extract class-discriminative representations from shape tokens, the adapter combines multi-resolution convolutional encoding with attention-based aggregation in a lightweight design. Its core consists of three parallel 1D CNNs with varying kernel sizes~\cite{ismail2020inceptiontime}, capturing discriminative temporal patterns across multiple resolutions. To aggregate these features into a class token, we employ an attention pooling mechanism with a linear complexity~\cite{ilse2018attention}. For each scale $q$, an attention head $\psi_{\text{ATTN}}$ assigns weights $\alpha_i^{(q)} \in [0,1]$ to shape tokens $\mathbf{z}_i^{(q)} \in \mathbb{R}^d$ via two linear layers with $\tanh$ and sigmoid activations. The aggregated class token $\mathbf{c}^{(q)}$ is computed as:
\begin{equation}
\alpha_i^{(q)} = \psi_{\text{ATTN}}(\mathbf{z}_i^{(q)}), \quad 
\mathbf{c}^{(q)} = \sum_{i=1}^{N_q} \alpha_i^{(q)} \mathbf{z}_i^{(q)},
\label{eq:shape_atten_fusion}
\end{equation}
where $\alpha_i^{(q)}$ reflects the discriminative importance of each shape, enhancing the interpretability of classification results.

To integrate information from multi-scale shape tokens, we adopt a coarse-to-fine (from larger to smaller $W_q$) hierarchical fusion strategy. At each scale $q > 1$, the previous class token $\mathbf{c}^{(q{-}1)}$ is prepended to the shape tokens:
\begin{equation}
\mathcal{Z}^{(q)} = \left[ \mathbf{c}^{(q{-}1)} \right] \oplus \left[ \mathbf{z}_i^{(q)} \right]_{i=1}^{N_q},
\label{eq:shape_adapter}
\end{equation}
which enables the hierarchical fusion of class tokens and aggregates discriminative patterns across shape scales.

After fusing the $Q$ input scales, we obtain a final class token $\mathbf{c}^{(Q)}$ and corresponding shape tokens $[\mathbf{z}_i^{(Q)}]_{i=1}^{N_Q}$. Each fixed-scale subsequence set $\mathcal{S}^{(q)}$ is independently normalized to produce shape tokens $[\mathbf{z}_i^{(q)}]_{i=1}^{N_q}$. The shape-aware adapter then applies attention pooling to map each token set to a $d$-dimensional class token $\mathbf{c}^{(q)}$. This adapter, shared across all scales $q \in [1, Q]$, serves as a unified module that adaptively highlights discriminative temporal features. The design supports efficient multi-scale integration and facilitates transferability of shapelet patterns during pretraining.

\subsection{Prototype-based Pretraining}

Prototype learning captures class-level features by learning embeddings for each class, enabling strong generalization and domain adaptation with limited labeled data~\cite{snell2017prototypical}.
Class labels play a vital role in guiding the learning of shapelet representations. This supervision helps the adapter to select the optimal temporal scales, enhancing hierarchical class token fusion. Notably, attention scores computed during shape token aggregation (Eq.~(\ref{eq:shape_atten_fusion})) can be ambiguous without class-level signals, making it difficult to assess the discriminative relevance of each shape token.
To address this, we introduce a prototype-based pretraining module that aligns class and shape tokens with their respective class prototypes, thereby reducing reliance on large amounts of source labeled data.
This is achieved through instance-level and shape-level contrastive objectives that encourage learning transferable shapelet patterns.

\subsubsection{Instance-Prototype Contrastive Learning.}

To improve token representations,  UniShape adopts a Transformer encoder as its backbone, given its demonstrated effectiveness in FM architectures~\cite{liang2024foundation}. 
The class token $\mathbf{c}^{(Q)}$ and associated shape tokens $[\mathbf{z}_i^{(Q)}]_{i=1}^{N_Q}$, produced by the shape-aware adapter, are fed into the transformer encoder to obtain refined outputs $\mathbf{c}^{(Q)\prime}$ and $[\mathbf{z_i}^{(Q)\prime}]_{i=1}^{N_Q}$.

Contrastive learning has proven effective in pretraining by encouraging the model to learn discriminative features through the comparison of positive and negative pairs~\cite{chen2020simple}. Motivated by this, we introduce an \textit{instance-level contrastive learning} strategy to align class tokens with their corresponding class prototypes. This approach is composed of two key components: (i) the initialization and optimization of a set of learnable class prototype vectors, and (ii) the formulation of a contrastive loss using the class tokens.

Specifically,
we define a learnable prototype set as:
\begin{equation}
\mathcal{P} = \{\mathbf{p}_1, \dots, \mathbf{p}_c, \dots, \mathbf{p}_C\}, \quad \mathbf{p}_c \in \mathbb{R}^d,
\end{equation}
where $C$ is the number of classes of the training dataset and $\mathbf{p}_c$ denotes the prototype embedding for class $c$. These prototypes are dynamically updated during training. For each labeled instance, the class token $\mathbf{c}^{(Q)\prime}$ is used to update its corresponding prototype via exponential moving average:
\begin{equation}
\mathbf{p}_y \leftarrow \beta \mathbf{p}_y +  (1 - \beta) \mathbf{c}^{(Q)\prime},
\end{equation}
where $y$ is the ground-truth class label and $\beta \in (0,1)$ is the momentum coefficient. 

For unlabeled samples, the class token $\mathbf{c}^{(Q)\prime}$ is assigned a pseudo-label by identifying the nearest class prototype $\mathbf{p}_+$ according to cosine similarity. The instance-prototype contrastive loss is defined as:
\begin{equation}
\mathcal{L}_{\text{ins}} = -\log \frac{ \exp(\text{sim}(\mathbf{c}^{(Q)\prime}, \mathbf{p}_+) / \tau) }{ \sum_{j=1}^{C} \exp(\text{sim}(\mathbf{c}^{(Q)\prime}, \mathbf{p}_j) / \tau) },
\end{equation}
where $\text{sim}(\cdot, \cdot)$ denotes cosine similarity and $\tau$ is a temperature scaling factor.
Using class tokens for prototype learning allows the model to capture global class-discriminative features across samples.

\subsubsection{Shape-Prototype Contrastive Learning.}

Due to potential intra-class distributional variance and the limited ability of class tokens to capture local discriminative features within the time series, relying solely on instance-level class tokens for prototype learning may hinder the acquisition of optimal shapelet representations. Therefore, we introduce \textit{shape-level contrastive learning} based on the shape tokens $[\mathbf{z}_i^{(Q)\prime}]_{i=1}^{N_Q}$. The attention head from the adapter (Eq.~(\ref{eq:shape_atten_fusion})) are reused to select high-confidence shape tokens. Let $\mathcal{Z}_{\text{top}} \subset [\mathbf{z}_i^{(Q)\prime}]_{i=1}^{N_Q}$ denote the top-$\epsilon$ tokens with the highest scores. The shape-prototype contrastive loss is defined as:
\begin{equation}
\resizebox{.90\hsize}{!}
{$\mathcal{L}_{\text{shape}} = \frac{1}{|\mathcal{Z}_{\text{top}}|} \sum_{\mathbf{z_i}^{(Q)\prime} \in \mathcal{Z}_{\text{top}}} 
-\log \frac{ \exp(\text{sim}(\mathbf{z_i}^{(Q)\prime}, \mathbf{p}_+) / \tau) }{ \sum_{j=1}^{C} \exp(\text{sim}(\mathbf{z_i}^{(Q)\prime}, \mathbf{p}_j) / \tau) }$,}
\end{equation}
where the prototype $\mathbf{p}_+$ for each shape token is determined either by the ground-truth label or the pseudo-label of the corresponding class token $\mathbf{c}^{(Q)\prime}$.

Hence, the prototype-based pretraining loss combines both levels of contrastive learning:

\begin{equation}
\mathcal{L}_{\text{proto}} = (1 - \lambda) \mathcal{L}_{\text{ins}} + \lambda \mathcal{L}_{\text{shape}},
\end{equation}
where $\lambda \in (0, 1)$ is a hyperparameter that balances the contributions of the instance-level and shape-level losses. 

\subsection{The Overall Training Loss Function}

\subsubsection{Pretraining Loss.}

To improve training stability and increase sample diversity, we adopt the momentum contrastive learning framework of MoCo v3~\cite{chen2021empirical} for pretraining. For each input time series $\mathbf{x}$, two augmented views $\mathbf{x}_1$ and $\mathbf{x}_2$ are generated via random cropping~\cite{linnutime2024} and independently encoded to produce class and shape tokens.
To support pretraining under weak supervision, we employ the MoCo v3 self-supervised contrastive loss, which enforces consistency between different views of the same instance. This objective promotes representation learning from large-scale unlabeled data and is defined as:
\begin{equation}
\mathcal{L}_{\text{self}} = - \log \frac{ \exp\left( \mathrm{sim}(\mathbf{q}, \mathbf{k}^+) / \tau \right) }{ \sum\limits_{j} \exp\left( \mathrm{sim}(\mathbf{q}, \mathbf{k}_j) / \tau \right) },
\end{equation}
where $\mathbf{q}$ is the query embedding, $\mathbf{k}^+$ is the positive key from the momentum encoder.

Hence,  the overall pretraining objective is formulated as:
\begin{equation}
\mathcal{L}_{\text{pretrain}} = \mathcal{L}_{\text{proto}} + \mathcal{L}_{\text{self}}.
\end{equation}

\subsubsection{Fine-tuning Loss.}

The UniShape model is fine-tuned on the target dataset using the pre-trained adapter, encoder and a randomly initialized classification head. Supervised training is conducted using the cross-entropy loss:
\begin{equation}
\mathcal{L}_{\text{ce}} = - \sum_{i=1}^{C} y_i \log(\hat{y}_i),
\end{equation}
where $\hat{y}_i$ denotes the predicted probability for class $i$ of sample $\mathbf{x}_t$, and $y_i$ is the one-hot encoded ground-truth label.

To enhance interpretability, we incorporate the $\mathcal{L}_{\text{shape}}$ as an auxiliary objective. This loss aligns shape tokens with their corresponding class labels, encouraging the model to focus on discriminative shapelet patterns. Therefore, the overall fine-tuning loss is defined as:
\begin{equation}
\mathcal{L}_{\text{finetune}} = \mathcal{L}_{\text{ce}} + \mu \mathcal{L}_{\text{shape}},
\end{equation}
where $\mu$ controls the training weight of $\mathcal{L}_{\text{shape}}$.

\section{Experiments}

\subsubsection{Datasets.}

UniShape is pre-trained on a source dataset of 1.89 million samples using five subsequence scales ($Q=5$), with window lengths and strides $W_q = K_q \in \{64, 32, 16, 8, 4\}$. The reason for these window length settings is detailed in Appendix~A. For downstream TSC tasks, UniShape is evaluated on 158 univariate datasets, including 128 UCR time series datasets widely used for classification tasks~\cite{dau2019ucr}. To evaluate zero-shot generalization, we further assess performance on 30 datasets from diverse domains~\cite{middlehurst2024bake}, which are not included in the pretraining source dataset. All datasets use their official train/test splits for evaluation.

\subsubsection{Baselines.}
We compare UniShape against 16 methods, grouped as follows:
(i) \textbf{Non-deep learning (NDL)}: Rocket~\cite{dempster2020rocket}, MiniRocket~\cite{dempster2021minirocket}, RDST~\cite{guillaume2022random}, MultiRocket-Hydra (MR-H)~\cite{dempster2023hydra};
(ii) \textbf{Domain-specific deep learning (DS)}: InceptionTime~\cite{ismail2020inceptiontime}, TS2Vec~\cite{yue2022ts2vec}, PatchTST~\cite{nietime2023}, TimesNet~\cite{wutimesnet2023}, SoftShape~\cite{liusoftshape2025};
(iii) \textbf{Foundation models (FMs)}: GPT4TS~\cite{zhou2023one}, MOMENT~\cite{goswami2024moment}, UniTS~\cite{gao2024units}, NuTime~\cite{linnutime2024}, Mantis~\cite{feofanov2025mantis}. We also include two forecasting-based FMs for zero-shot classification: Chronos~\cite{ansari2024} and Moirai~\cite{woo2024unified}.
All baselines use author-recommended hyperparameters in a unified Python environment for fair comparison.

\subsubsection{Implementation Settings.}

UniShape training comprises two stages: pretraining and fine-tuning. 
Pretraining is conducted for up to 30 epochs with a batch size of 2048. By default, prototype-based pretraining uses 10\% labeled data, a momentum coefficient $\beta = 0.9$, shape token selection ratio $\epsilon = 60\%$, and a weighting factor $\lambda = 0.01$. Fine-tuning runs for up to 300 epochs with an auxiliary loss weight $\mu = 0.01$. 
For deep learning-based methods, we follow the settings in~\cite{early2024inherently} and report test classification accuracy using the model checkpoint with the lowest training loss. For all baselines and UniShape, results are averaged over five independent runs with different random seeds. All experiments are conducted using Python 3.9.21, PyTorch 1.12.1, and four NVIDIA RTX A6000 GPUs.

Further details on the datasets, baselines, and implementation settings are provided in Appendix A and available at~\url{https://github.com/qianlima-lab/UniShape}.

\begin{table}[]
\centering
\resizebox{\linewidth}{!}{
\begin{tabular}{@{}c|l|c|ccc@{}}
\toprule
 & Method & \# Params & Avg. Acc & Avg. Rank & P-value \\ 
\midrule
\multirow{4}{*}{\rotatebox[origin=c]{90}{NDL}} 
 & Rocket & - & 0.8487 & 4.87 & 7.80E-06 \\
 & MiniRocket & - & 0.8545 & 4.84 & 3.25E-03 \\
 & RDST & - & 0.8571 & 4.80 & 8.58E-03 \\
 & MR-H & - & 0.8621 & 3.97 & 2.96E-02 \\ 
\noalign{\vskip 1pt} \hline \noalign{\vskip 1pt}
\multirow{5}{*}{\rotatebox[origin=c]{90}{DS}}
 & InceptionTime & 386.9 K & 0.8315 & 6.10 & 2.55E-11 \\
 & TS2Vec & 637.2 K & 0.8016 & 8.32 & 3.12E-11 \\
 & PatchTST & 431.2 K & 0.6500 & 12.69 & 2.12E-26 \\
 & TimesNet & 7.4 M & 0.6897 & 11.83 & 3.08E-24 \\
 & SoftShape & 472.5 K & 0.8388 & 5.89 & 3.68E-32 \\ 
\noalign{\vskip 1pt} \hline \noalign{\vskip 1pt}
\multirow{6}{*}{\rotatebox[origin=c]{90}{FMs}}
 & GPT4TS$^*$ & 84.1 M & 0.7100 & 11.69 & 7.62E-23 \\
 & MOMENT$^*$ & 341.2 M & 0.7020 & 12.10 & 7.04E-25 \\
 & UniTS & 1.1 M & 0.7357 & 11.43 & 3.20E-23 \\
 & NuTime & 2.4 M & 0.8353 & 6.68 & 2.08E-10 \\
 & Mantis & 8.7 M & 0.8441 & 5.21 & 1.69E-06 \\
 \cmidrule(l){2-6}  
 & \textbf{UniShape} & 3.1 M & \textbf{0.8708} & \textbf{2.71} & - \\ 
\bottomrule
\end{tabular}}
\caption{The statistical test results comparison on 128 UCR datasets under the fully supervised setting. 
\textit{\# Params} denotes the number of parameters in baselines. $^*$ indicates that GPT4TS and MOMENT utilize only 2.9 million (M) and 3.1 thousand (K) parameters for fine-tuning, respectively. Best results are in \textbf{bold}.
}
\label{tab:main_table_1}
\end{table}

\subsection{Overall Evaluation Results}

Table~\ref{tab:main_table_1} presents the test classification performance of UniShape compared to baselines on 128 UCR datasets under a fully supervised setting. Detailed per-dataset results are provided in Appendix B. Each method is evaluated using three metrics: \textit{Avg. Acc} (average accuracy across 128 datasets), \textit{Avg. Rank} (test accuracy-based rank), and P-value (Wilcoxon signed-rank test~\cite{demvsar2006statistical} assessing whether UniShape significantly outperforms each baseline; significance at $p <$ 0.05). Parameter counts exclude NDL-based methods due to their non-trainable architectures.

UniShape achieves the highest \textit{Avg. Acc} and the lowest \textit{Avg. Rank}, indicating strong generalization across diverse target datasets. GPT4TS uses a pre-trained language model for fine-tuning, while MOMENT and UniTS lack design considerations for TSC tasks, resulting in substantially lower accuracy than most NDL and DS-based methods. This suggests that task-agnostic FMs not specifically designed for TSC may struggle to adapt effectively to the classification task.
In contrast, UniShape significantly outperforms other FMs such as NuTime and Mantis, despite having only 3.1 million parameters. These findings collectively demonstrate that UniShape provides a more effective and parameter-efficient foundation for TSC.

\subsection{Pretraining Analysis}

\begin{figure}
	\centering 
	\includegraphics[ width=0.47\textwidth]{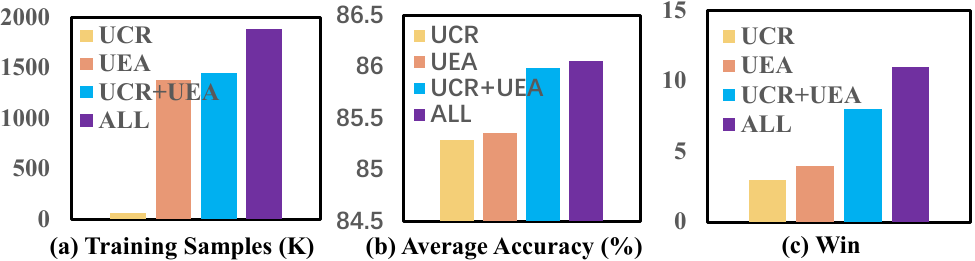}
	\caption{
    Results comparison on 18 UCR datasets with different numbers of pretraining samples. \textit{Win} denotes the number of datasets where the method performs best.
    } 
	\label{fig:ana_data_size}
\end{figure}

We analyze the impact of pretraining on two dimensions: the size of the pretraining dataset and the labeling ratio of its samples.
Following~\cite{liusoftshape2025}, and considering the computational cost of using all 128 UCR datasets as target domains, we select 18 UCR datasets that vary in domain, class count, and sample size for analysis.
Figure~\ref{fig:ana_data_size} illustrates UniShape’s performance across four pretraining scales: UCR ($\sim$60K), UEA ($\sim$1.39M), UCR+UEA ($\sim$1.45M), and ALL (1.89M samples). Classification accuracy consistently improves with larger pretraining datasets, indicating that data scale enhances target-domain performance, while pretraining solely on the smaller UCR archive still yields competitive fine-tuning results.

\begin{table}[]
\centering
\resizebox{\linewidth}{!}{
\begin{tabular}{@{}l|ccccc@{}}
\toprule
Labeling Ratio & 0\% & 1\% & 10\% & 50\% & 100\% \\ \midrule
Avg. Acc & 0.8395 & 0.8410 & 0.8529 & 0.8574 & \textbf{0.8588} \\
P-value & 1.87E-02 & 3.28E-02 & 2.00E-01 & 4.10E-01 & - \\ \bottomrule
\end{tabular}}
\caption{The statistical test classification results comparison 
on 18 UCR datasets with different training sample labeling ratios for pretraining.  The best result is in \textbf{bold}.}
\label{tab:my-table-hyp-labeling-ratio}
\end{table}

Pretraining on the full 1.89 million samples is time-consuming per run, making it computationally intensive across all labeling ratios. For data scale analysis results in Figure~\ref{fig:ana_data_size}, we employ only the training samples from the UCR archive for pretraining to reduce computational cost.
In Table~\ref{tab:my-table-hyp-labeling-ratio}, the 0\% labeling ratio initializes prototypes randomly and updates them via pseudo-labels during pretraining. The 100\% labeled setting achieves the highest accuracy, demonstrating the effectiveness of incorporating supervised class information for fine-tuning. However, performance gains beyond 10\% labeling are marginal and statistically insignificant (P-value~$>$~0.05). Hence, we adopt 10\% labeling as the default setting for pretraining to balance performance and annotation cost. Detailed results of Figure~\ref{fig:ana_data_size} and Table~\ref{tab:my-table-hyp-labeling-ratio}, please refer to Appendix C.

\subsection{Interpretability  Analysis}

\begin{figure}
	\centering 
	\includegraphics[ width=0.47\textwidth]{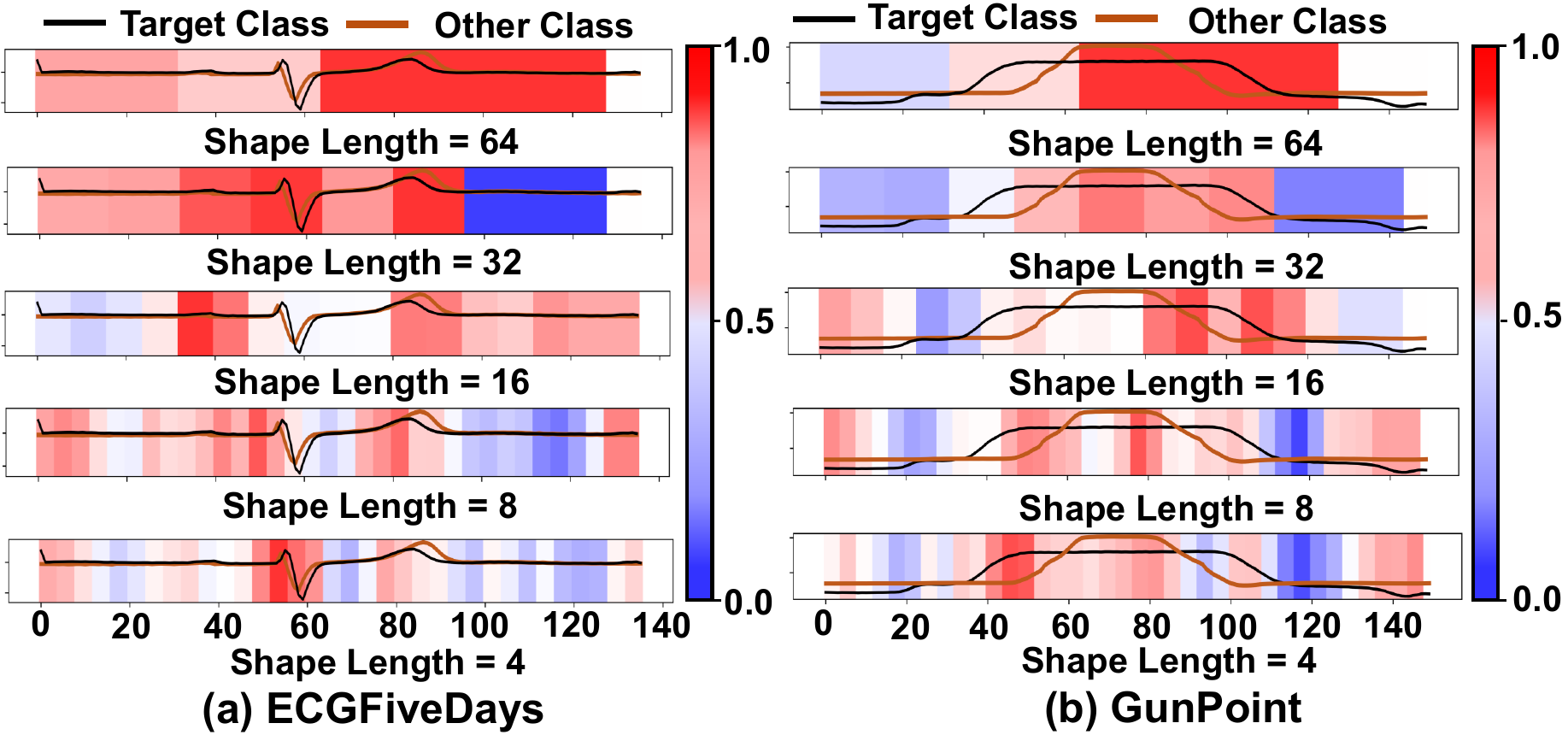}
	\caption{
   Visualization of attention scores learned by the shape-aware adapter across different shape lengths. Darker red denotes higher attention, highlighting discriminative regions for the target class, while darker blue indicates lower attention and reduced relevance to target features.
    } 
	\label{fig:inter_ana_vis}
\end{figure}

To evaluate the interpretability of UniShape, we examine the attention scores produced by the shape-aware adapter, specifically the attention head $\psi_{\text{ATTN}}$, across multiple shape lengths. As shown in Figure~\ref{fig:inter_ana_vis}, we conduct this analysis on two representative UCR datasets: \textit{ECGFiveDays} and \textit{GunPoint}, drawn from the healthcare and motion domains, respectively. In \textit{ECGFiveDays},~\citet{rakthanmanon2013fast} identified the delayed \textit{T-wave} in the [75, 95] interval as the key discriminative segment. UniShape consistently allocates high attention to this region, as well as to some regions with minimal temporal differences. For \textit{GunPoint}~\cite{ye2011time}, which differentiates \textit{Gun} and \textit{NoGun} gestures, critical segments lie in [30, 60] and [90, 110] due to motion overshoot. 
UniShape effectively assigns high attention to these intervals, selecting them as shapelets to enhance classification interpretability. These results highlight UniShape’s advantages over existing time series FMs, which often neglect the interpretability of classification results.

\subsection{Results on Zero-Shot Feature Extraction}

\begin{table}[h]
\centering
{
\footnotesize
\begin{tabular}{l|ccc}
\toprule
Method & Avg. Acc & Avg. Rank & P-value \\
\midrule
RandomForest       & 0.6930 & 3.77 & 2.57E-02 \\
GPT4TS             & 0.5600 & 6.37 & 1.79E-06 \\
MOMENT             & 0.6972 & 4.17 & 3.98E-02 \\
Chronos             & 0.6793 & 4.10 & 4.91E-03 \\
Moirai             & 0.5691 & 6.37 & 5.69E-06 \\
UniTS              & 0.3431 & 8.90 & 6.24E-10 \\
NuTime             & 0.6917 & 3.53 & 3.36E-03 \\
Mantis             & 0.7052 & 3.67 & 3.15E-02 \\
\cmidrule(l){1-4} 
\textbf{UniShape}  & \textbf{0.7262} & \textbf{3.07} & - \\
\bottomrule
\end{tabular}}
\caption{The statistical test results comparison on 30 additional datasets with a zero-shot feature extraction. Best results are in \textbf{bold}.
}
\label{tab:method_comparison}
\end{table}

Zero-shot learning evaluates a model’s ability to generalize to unseen data distributions without fine-tuning on the target domain~\cite{pourpanah2022review}, providing a critical measure of the generalization ability of FMs.
Table~\ref{tab:method_comparison} presents the zero-shot feature extraction performance of UniShape and seven time series FMs on 30 additional time series datasets from diverse domains. For detailed results of Table~\ref{tab:method_comparison}, please refer to Appendix D. 
Following the setup in~\cite{feofanov2025mantis}, each dataset is processed using a frozen FM to extract representations for both training and test sets. A Random Forest classifier~\cite{breiman2001random} is then trained on these representations to assess classification performance. A baseline using Random Forest trained directly on raw time series is also included for comparison.

As shown in Table~\ref{tab:method_comparison}, UniShape outperforms all baselines. Mantis and NuTime also perform competitively, with better \textit{Avg. Rank} than the  RandomForest baseline. In contrast, GPT4TS, Moirai, and UniTS perform poorly for TSC in the zero-shot setting, suggesting a limited ability to capture classification-specific temporal patterns. These results demonstrate that UniShape exhibits strong generalization and transferability in zero-shot scenarios, underscoring its potential as a universal time series FM to extract meaningful class-discriminative shapelet features without fine-tuning.

\subsection{Ablation Study}

This section conducts a comprehensive ablation study examining the training paradigm and model architectures. Following the setup in Table~\ref{tab:my-table-hyp-labeling-ratio}, we only use the UCR archive for pretraining and the same 18 UCR datasets as target domains to reduce
computational cost.
Two training paradigms are evaluated:
(a) \textbf{w/o Pretraining}: models trained from scratch on target datasets.
(b) \textbf{Pretraining and Fine-tuning}: models are first pre-trained and then fine-tuned for classification.
The model architecture ablation includes three components:
(a) \textbf{Adapter Module}: i) \textit{w/o Adapter}: removes the shape-aware adapter, using a fixed-length subsequence set as input. ii) \textit{re Trans}: replaces the CNN layer in the adapter with a Transformer. iii) \textit{re MLP}: replaces the CNN layer with an MLP. (b) \textbf{Prototype-based pretraining}: i) \textit{w/o Ins}: removes the instance-prototype contrastive loss. ii) \textit{w/o Shape}: removes the shape-prototype contrastive loss. iii) \textit{w/o Proto}: removes both instance and shape prototype losses.
(c) \textbf{Transformer Encoder}: i) \textit{re CNN}: replaces the Transformer encoder with an Inception-based CNN. ii) \textit{re MLP}: replaces the Transformer encoder with an MLP.

\begin{table}[htbp]
\centering
\small %
\setlength{\tabcolsep}{3pt} 
\renewcommand{\arraystretch}{1.1} 
\begin{tabular}{@{}l|l|ccc@{}}
\toprule
\multicolumn{2}{c|}{\textbf{Method}} & Avg. Acc & Avg. Rank & P-value \\
\midrule
\multicolumn{2}{c|}{\textbf{UniShape}} & \textbf{0.8529} & \textbf{3.00} & -- \\
\midrule
\multicolumn{5}{l}{\textit{w/o Pretraining}} \\
\midrule
 & using Trans       & 0.8365 (-1.64\%) & 6.33 & 2.96E-02 \\
Encoder & re CNN            & 0.8264 (-2.65\%) & 6.61 & 2.33E-02 \\
 & re MLP            & 0.6832 (-17.0\%) & 8.11 & 3.35E-03 \\
\midrule
\multicolumn{5}{l}{\textit{Pretraining and Fine-tuning}} \\
\midrule
   & w/o Adapter      & 0.8428 (-1.01\%) & 5.50 & 1.65E-02 \\
Adapter   & re Trans         & 0.8446 (-0.83\%) & 5.72 & 1.68E-02 \\
   & re MLP           & 0.8431 (-0.98\%) & 6.78 & 2.70E-02 \\
\midrule
 & w/o Ins          & 0.8444 (-0.85\%) & 5.56 & 5.92E-03 \\
Prototype & w/o Shape        & 0.8470 (-0.59\%) & 4.39 & 2.11E-02 \\
 & w/o Proto & 0.8411 (-1.18\%) & 5.89 & 8.15E-03 \\
\midrule
\multirow{2}{*}{Encoder} 
& re CNN & {0.8512 (-0.16\%)} & {3.94} & 3.40E-01 \\
& re MLP & 0.5651 (-28.8\%) & 9.61 & 4.11E-04 \\
\bottomrule
\end{tabular}
\caption{Ablation results on 18 UCR time series datasets. 
Among them, \textit{w/o} means without, and \textit{re} means replace.
}
\label{tab:aba_ana_model}
\end{table}

Table~\ref{tab:aba_ana_model} reports the statistical results across all ablation settings. Detailed results for Table~\ref{tab:aba_ana_model} are provided in Appendix E.
Excluding the pretraining phase leads to a substantial performance decline, underscoring the importance of pretraining. Substituting the CNN layer in the adapter with a Transformer or MLP further reduces performance, indicating that CNNs are more effective for multi-scale shapelet learning.
In the prototype-based pretraining module, \textit{w/o Ins} yields lower performance than \textit{w/o Shape}, suggesting that instance-prototype pretraining is more crucial for shapelet representation learning. Removing both components (\textit{w/o Proto}) further degrades performance, confirming their complementary roles.
Previous work~\cite{eldele2024tslanet} has shown that CNNs outperform Transformers in a domain-specific training setting for TSC tasks. However, under the FM setting with pretraining and fine-tuning, the Transformer encoder achieves a better \textit{Avg. Rank} than the CNN-based encoder. In contrast, replacing the Transformer with an MLP as the encoder leads to poor classification performance, even worse than training from scratch. This indicates that MLPs struggle to learn generalizable shapelet representations for TSC through pretraining, compared to Transformers and CNNs. Further analysis of hyperparameters $\epsilon$, $\lambda$, $\mu$, and runtime is presented in Appendix~F.

\section{Conclusion}

We present a unified shape-aware foundation model named UniShape for time series classification. UniShape integrates a shape-aware adapter and a prototype-based pretraining module, enabling effective learning of multi-scale shapelet representations. 
By pretraining on a large-scale source dataset, UniShape captures transferable shapelet patterns applicable to diverse target domains.
Experiments show that UniShape outperforms baselines in both fully supervised and zero-shot learning settings. 
Yet, this work focuses solely on univariate TSC. In the future, we aim to extend UniShape toward modeling multivariate dependencies for more generalizable foundation models.

\section{Acknowledgments}

We thank the anonymous reviewers for their helpful feedbacks.
We thank Professor Eamonn Keogh and all the people who have contributed to the UCR \& UEA archives and other time series datasets. 
The work described in this paper was partially funded by the National Natural Science Foundation of China (Grant Nos. 62272173, 62273109), the Natural Science Foundation of Guangdong Province (Grant Nos. 2024A1515010089, 2022A1515010179), the Science and Technology Planning Project of Guangdong Province (Grant No. 2023A0505050106), the National Key R\&D Program of China (Grant No. 2023YFA1011601), the MTI WS Fund - AI for Manufacturing COE - Common Model Projects (Grant No. W24MCMF012),
and the China Scholarship Council program (Grant No. 202406150081).

\bibliography{aaai2026}

\end{document}